\documentclass[letterpaper, 10 pt, conference]{ieeeconf}  

\IEEEoverridecommandlockouts       
\usepackage{graphicx}
\usepackage{amsmath}
\usepackage{amssymb}
\usepackage{booktabs}
\usepackage{lipsum}  
\usepackage{subcaption}
\usepackage{tikz}
\usepackage{pifont}
\usepackage{tcolorbox}
\usepackage[pagebackref,breaklinks,colorlinks]{hyperref}

\newcommand{\figTeaser}{
\begin{figure}
    \centering
    \includegraphics[width=\linewidth]{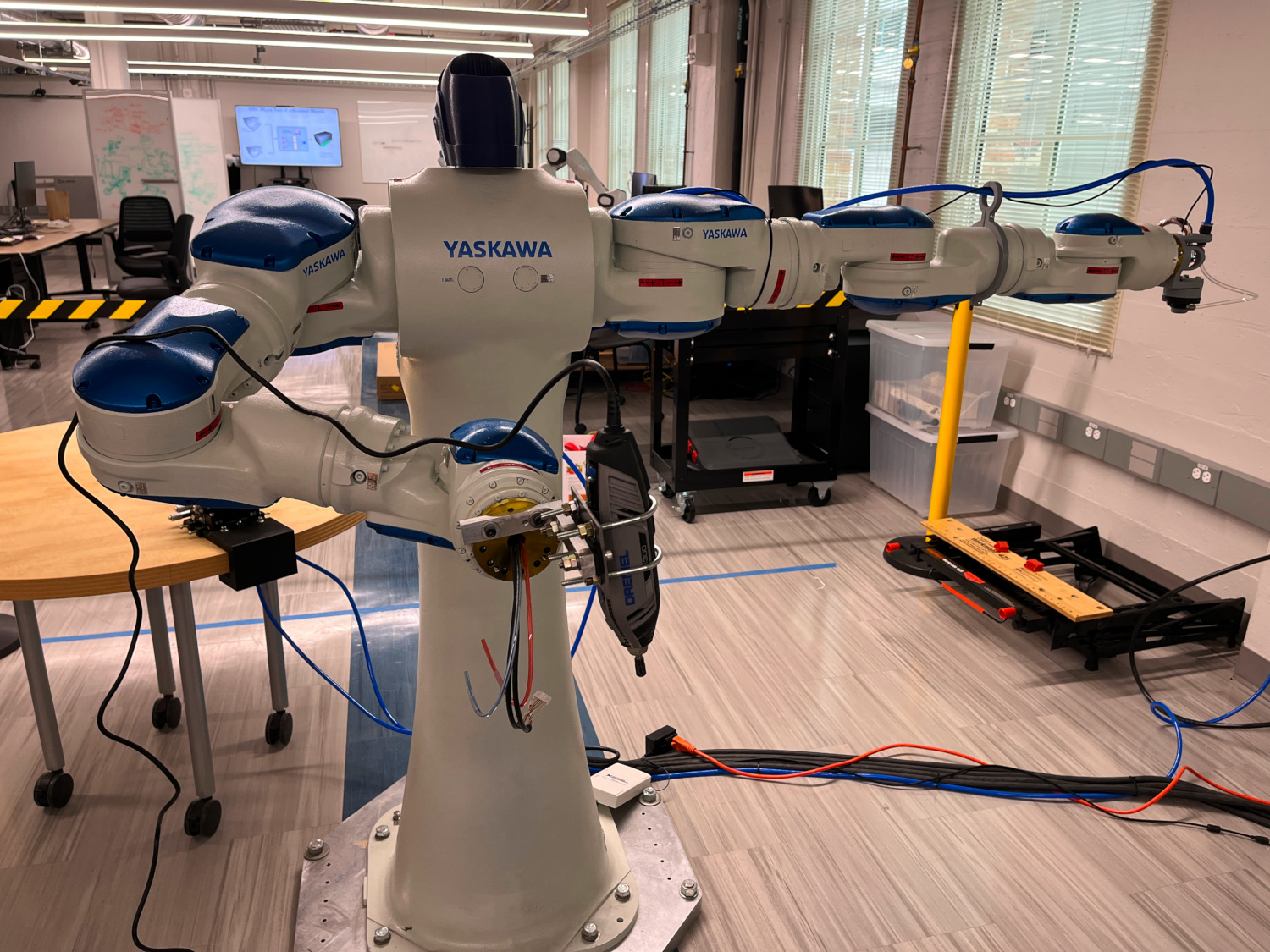}
    \caption{\small \textbf{Robot for synthetic data generation.} An industry-grade Yaskawa robot arm equipped with the smart hand tool at its end-effector. Robot arms like this offer a potential solution to the massive data demands inherent in machine learning tasks applied to physical applications, such as mechanical engineering.}
    \vspace{-5mm}
    \label{fig:teaser}
\end{figure}
}

\newcommand{\figSysArch}{
\begin{figure}[!htbp]
\centering
\includegraphics[width=\linewidth]{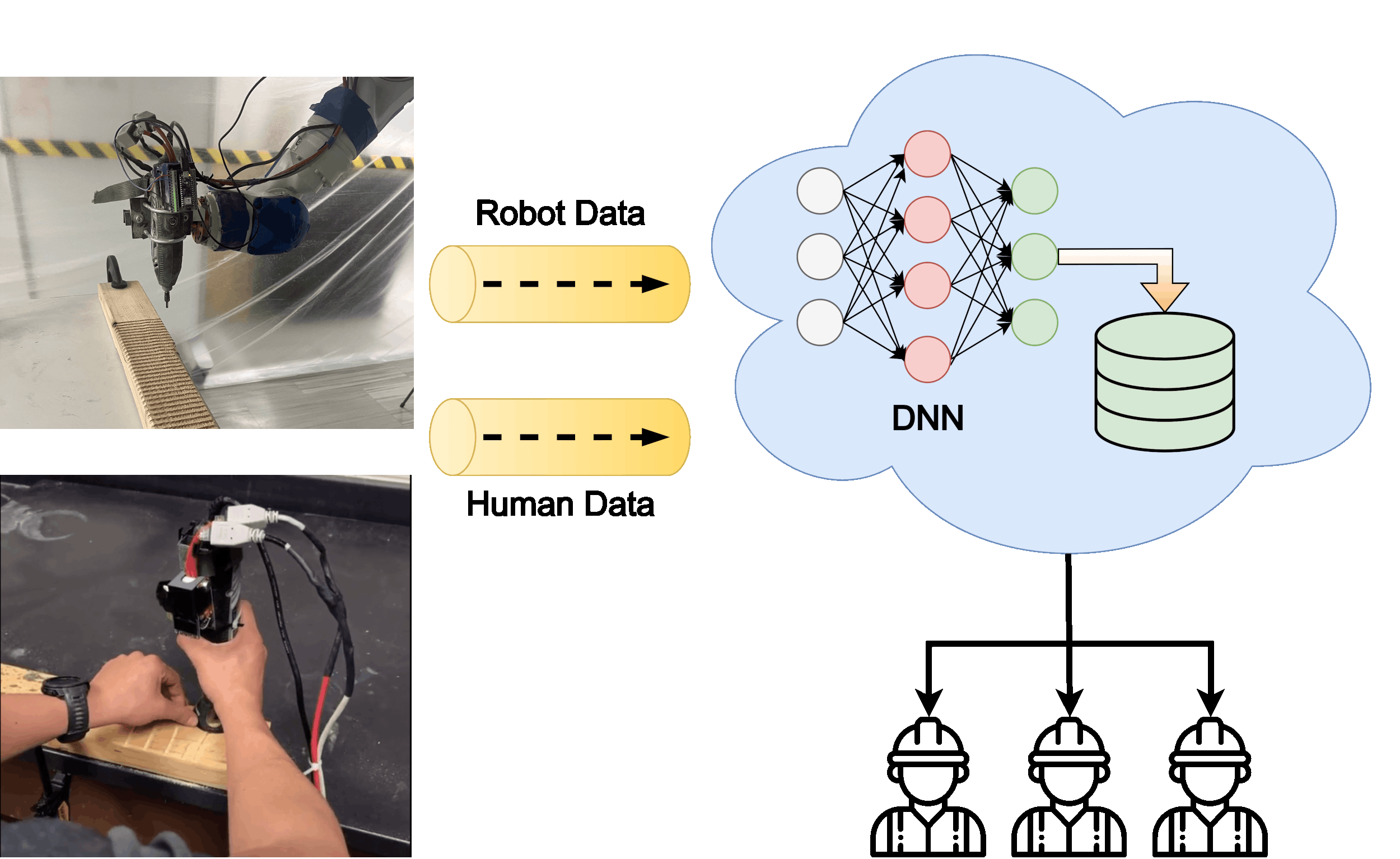}
\caption{\small \textbf{System architecture.} Data from multiple sensors is collected separately from both human subjects and a Yaskawa robot. Initially, the robot-collected data is used for pre-training a model. Later, the same model is fine-tuned on human-collected data on a per-subject basis, for better individual performance. The significance of this architecture lies in its scalability: pre-training robot-collected data can improve performance for increasing number of subjects and tasks.}
\label{fig:SysArch}
\end{figure}
}

\newcommand{\figSTM}{
\begin{figure}[!htbp]
\centering
\includegraphics[width=\linewidth]{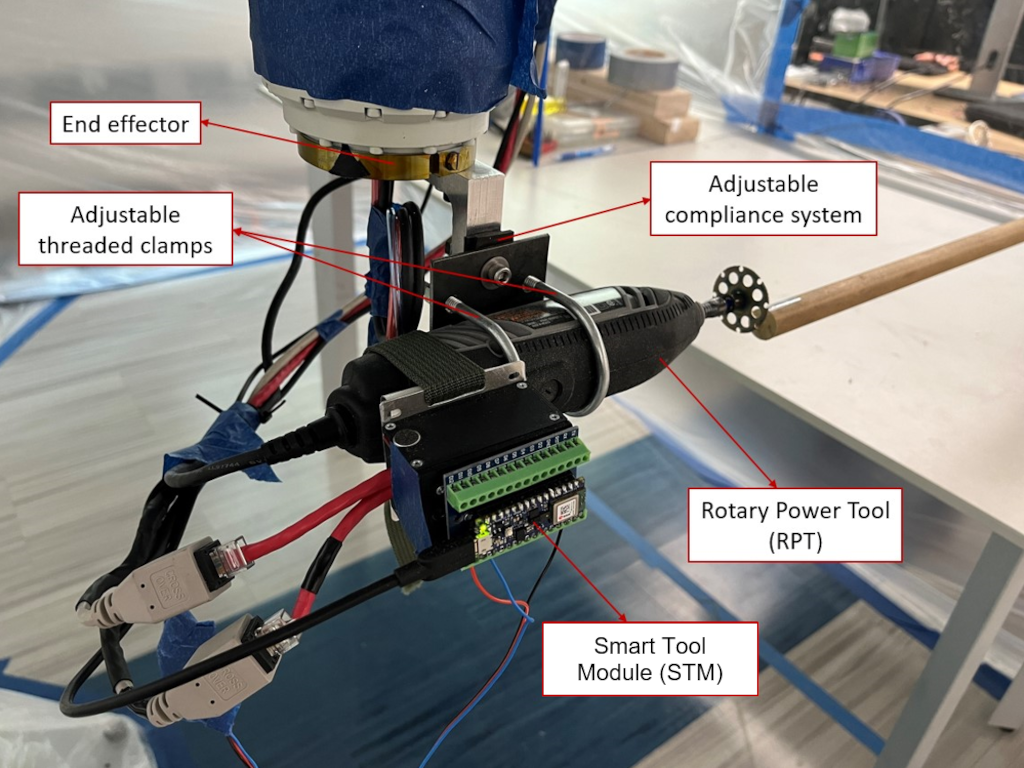}
\caption{\small \textbf{Yaskawa robot arm with the smart hand tool attached.} The robot operates within an enclosed area to prevent the dispersion of dust generated during the tool's operation. Different parts of the tool are labeled in the figure. In this experiment, the smart hand tool is equipped with a wood-cutting wheel, which is employed for the purpose of slicing a cylindrical wooden rod (see Section~\ref{sec:DataCollection}). A simple setup like this is able to mimic humans performing the tasks (cutting in this case) in the real world.}
\label{fig:STM}
\end{figure}
}

\newcommand{\figToolBits}{
\begin{figure}[!htbp]
\centering
\includegraphics[width=\linewidth]{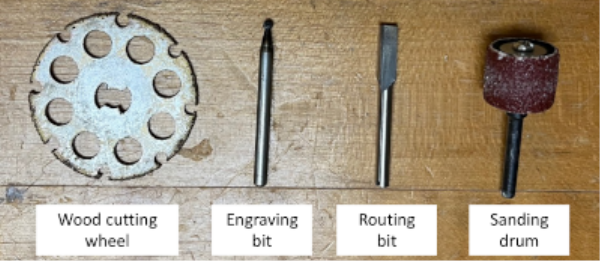}
\caption{\small \textbf{Tool bits.} From left to right: A wood cutting wheel, an engraving bit, a routing bit, and a sanding drum. Each bit allows the worker to perform a different activity associated with it. These bits were attached to the RPT as shown in Figure~\ref{fig:STM}.}
\label{fig:ToolBits}
\end{figure}
}

\newcommand{\figRawSensorDataRouting}{
\begin{figure}[!htbp]
     \centering
     \begin{subfigure}[b]{\linewidth}
         \centering
         \includegraphics[width=\textwidth]{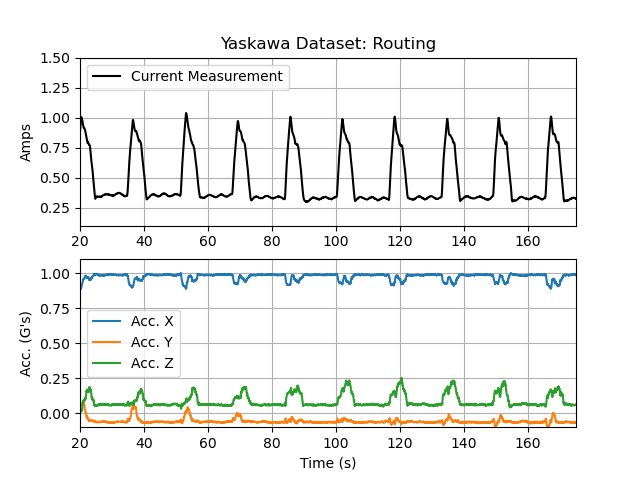}
     \end{subfigure}
     \begin{subfigure}[b]{\linewidth}
         \centering
         \includegraphics[width=\textwidth]{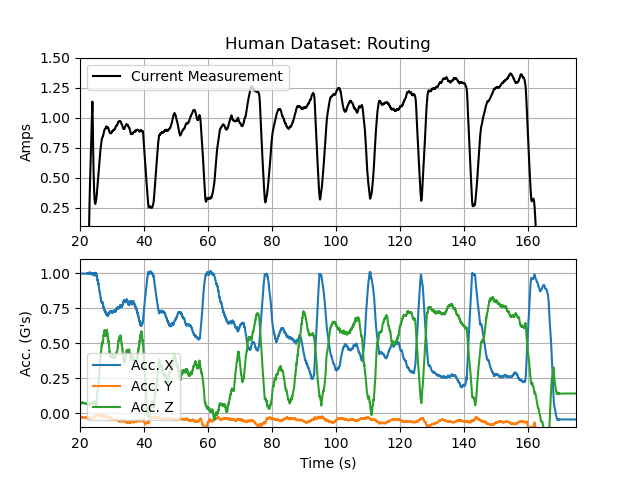}
     \end{subfigure}
     \caption{\small \textbf{Typical current draw and accelerations during a routing task.} The robot data (top) reflects a much more consistent motion compared to human data (bottom) due to differences in tool usage. Similar patterns were seen for similar tasks, but distinct across task types. This alludes to a pre-trained approach in the ML pipeline.}
     \label{fig:RawDataRouting}
\end{figure}
}

\newcommand{\figSensorComparisons}{
\begin{figure}[!htbp]
\centering
\includegraphics[width=\linewidth]{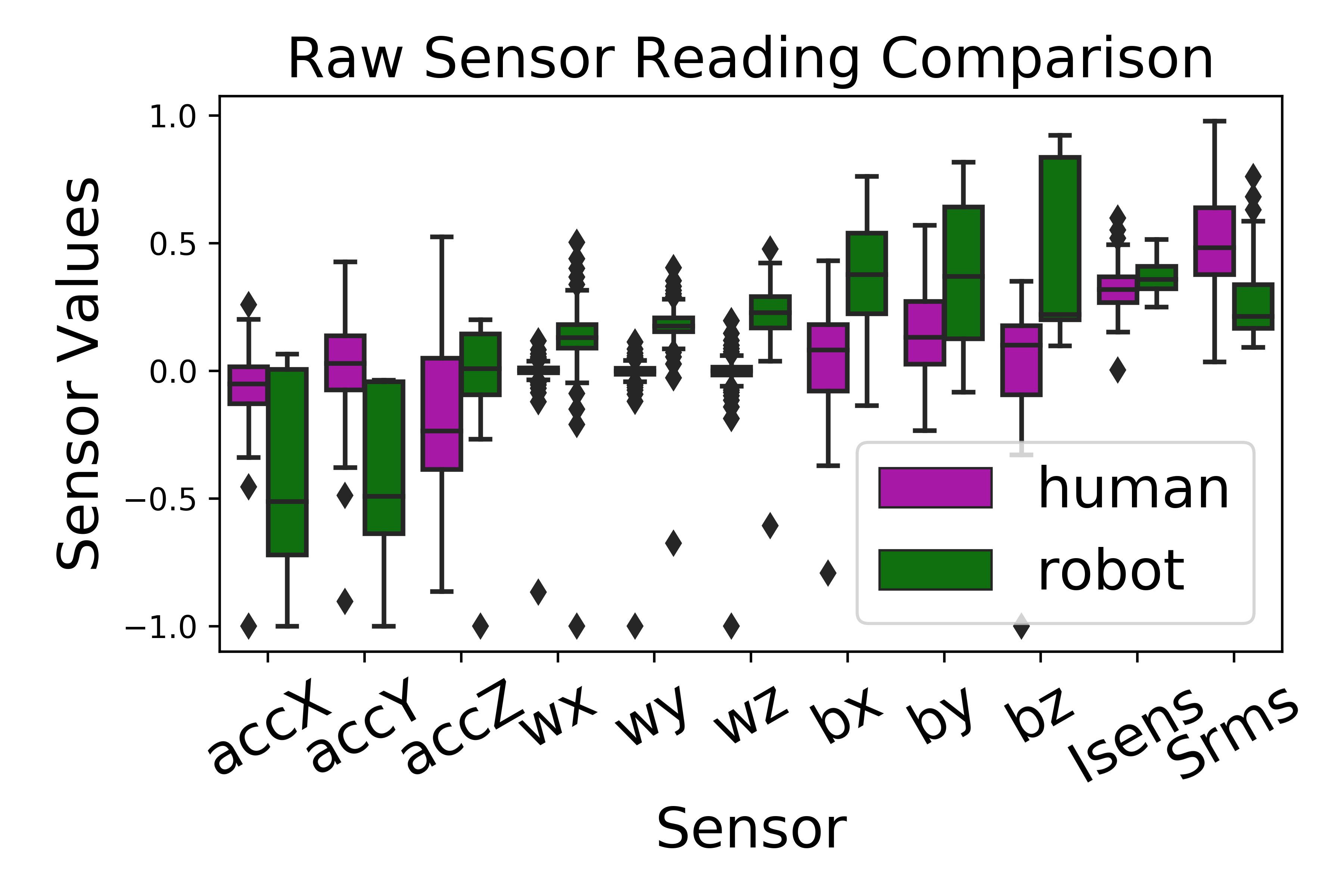}
\caption{\small \textbf{Grouped boxplot comparing raw sensor values.} The x-axis is each sensor on the STM attached to the RPT. The y-axis is the range of values normalized from $(-1,1)$ to account for different scales in different sensors. The diamonds in each x-axis entry represent the outliers of that distribution. The similarity of data distributions warrant the potential use of robot data for pre-training models.}
\label{fig:SensorComparisons}
\end{figure}
}

\newcommand{\figZSPreTrain}{
\begin{figure}[!htbp]
     \centering
     \includegraphics[width=\linewidth]{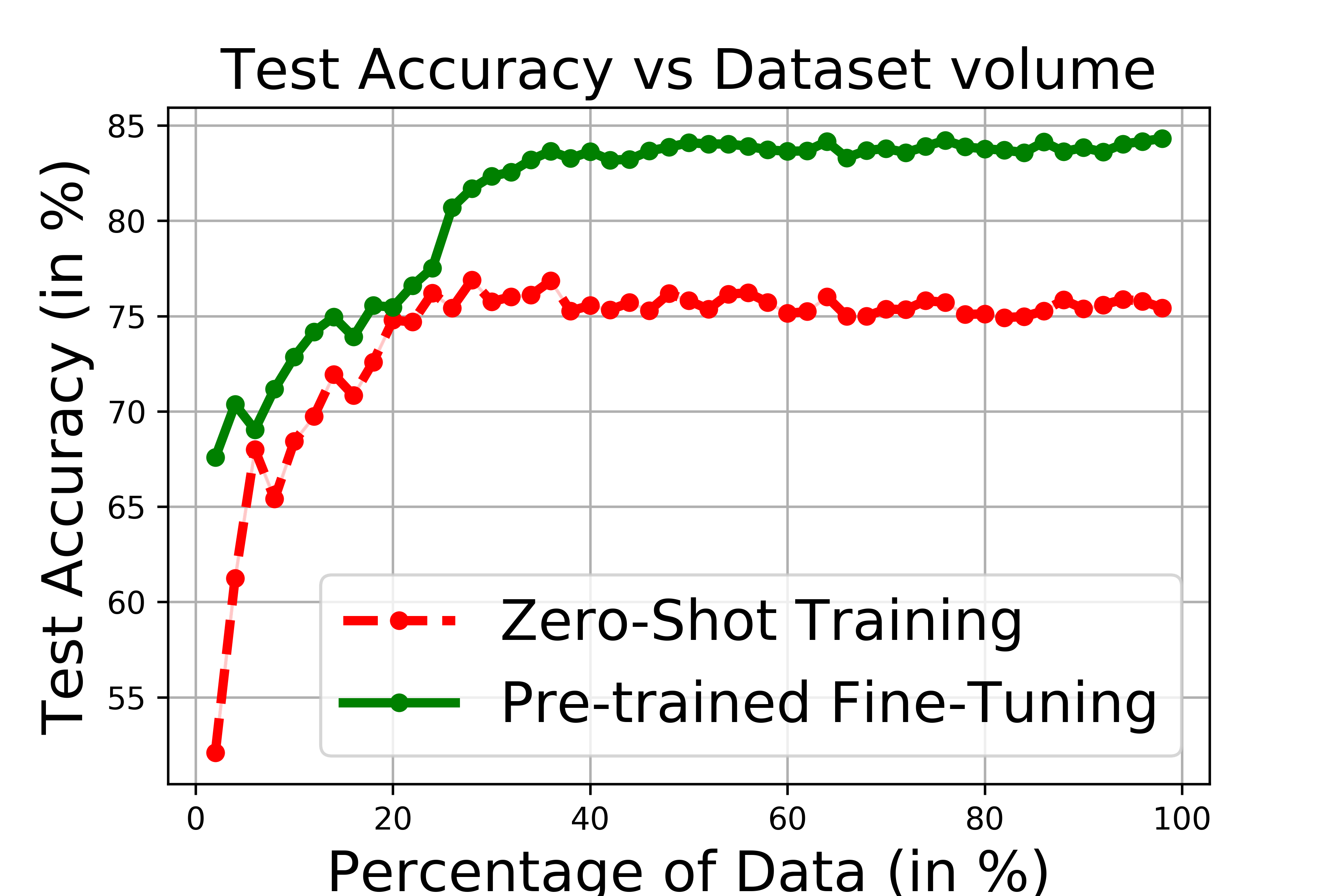}
     \caption{\small \textbf{Test accuracy versus dataset volume.} Here we see the zero-shot training on all human subject data (in \textbf{red}), versus fine-tuning a pre-trained model on the synthetic robot-collected data (in \textbf{green}). In the plot, we measure the accuracy on an unseen test dataset (higher is better). Clearly, these trends follow the law of diminishing returns in ML. Moreover, the fine-tuning approach outperforms the zero-shot training by a quantifiable margin.}
     \label{fig:ZSPretrainAllHumans}
\end{figure}
}

\newcommand{\figZSPreTrainSubjects}{
     \begin{figure}[!htbp]
         \centering
         \includegraphics[width=\linewidth]{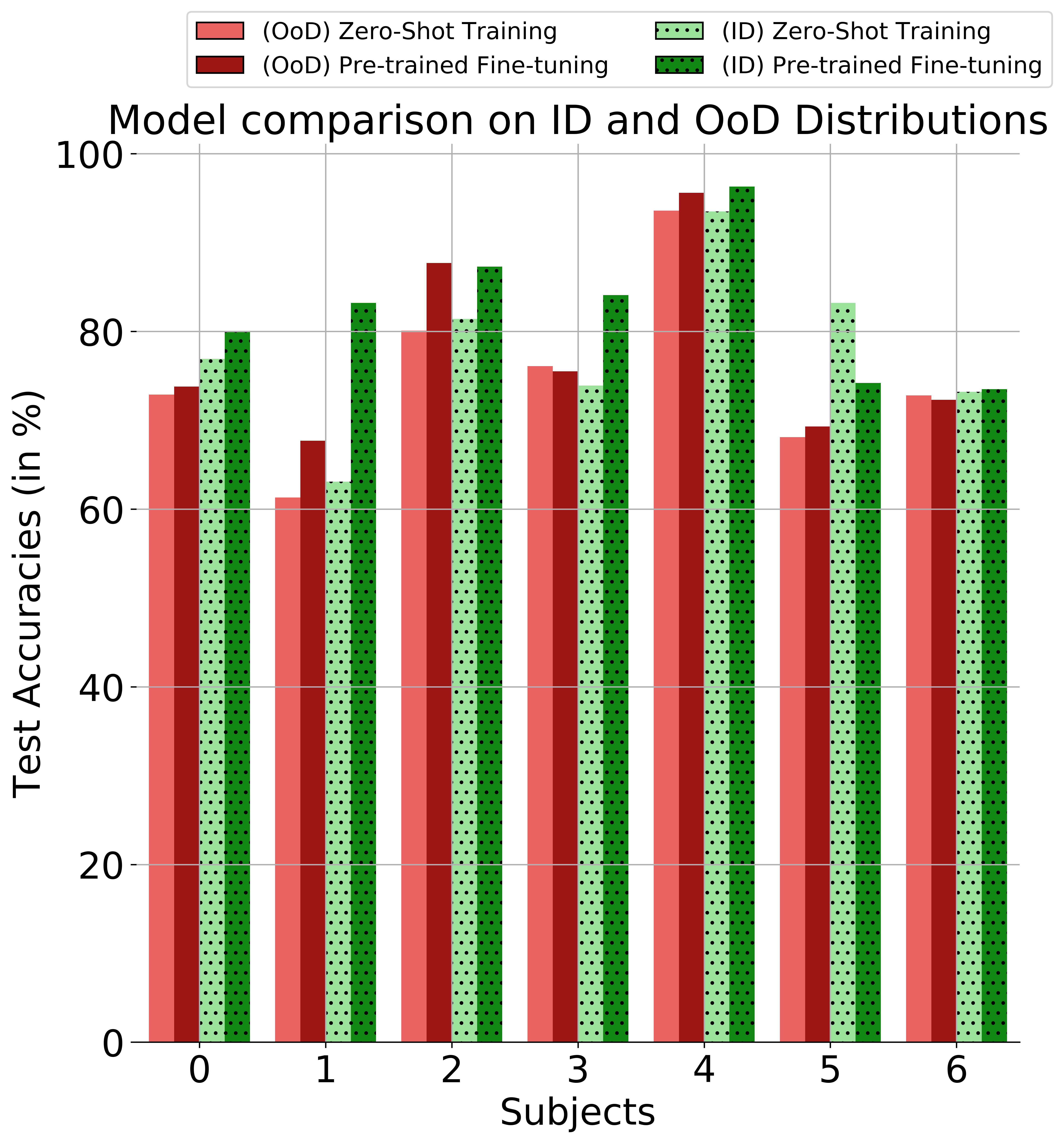}
         \caption{\small \textbf{Model performance on individual subjects.} This grouped barplot describes the differences in model performance on in-distribution (ID) versus out-of-distribution (OoD) test data. The red bars for each subject denote that fine-tuning on a pre-trained model is generally beneficial and boosts individual local model performance. }
     \label{fig:ZSPreTrainSubjects}
     \end{figure}
}

\usepackage{comment}
\usepackage{multirow}

\title{\LARGE \bf
Using human and robot synthetic data for training smart hand tools
}

\author{Jose Benda\~{n}a$^{1}$, Sundar Sripada V. S.$^{2}$, Carlos D. Salazar$^{1}$, Sandeep Chinchali$^{2}$ and Raul G. Longoria$^{1}$
\thanks{$^{1}$Department of Mechanical Engineering, $^{2}$Department of Electrical and Computer Engineering, The University of Texas at Austin}%
}
\begin{document}

\maketitle
\thispagestyle{empty}
\pagestyle{empty}

\begin{abstract}


The future of work does not require a choice between human and robot. Aside from explicit human-robot collaboration, robotics can play an increasingly important role in helping train workers as well as the tools they may use, especially in complex tasks that may be difficult to automate or effectively roboticize. This paper introduces a form of smart tool for use by human workers and shows how training the tool for task recognition, one of the key requirements, can be accomplished. Machine learning (ML) with purely human-based data can be extremely laborious and time-consuming. First, we show how data synthetically-generated by a robot can be leveraged in the ML training process. Later, we demonstrate how fine-tuning ML models for individual physical tasks and workers can significantly scale up the benefits of using ML to provide this feedback. Experimental results show the effectiveness and scalability of our approach, as we test data size versus accuracy. Smart hand tools of the type introduced here can provide insights and real-time analytics on efficient and safe tool usage and operation, thereby enhancing human participation and skill in a wide range of work environments. Using robotic platforms to help train smart tools will be essential, particularly given the diverse types of applications for which smart hand tools are envisioned for human use.

\end{abstract}

\section{Introduction}

Machine learning (ML) for enabling devices and systems to support humans in the future of human work is poised for significant growth in the upcoming decades. This holds true especially for collaborative robots, and recently for smart hand tools designed to monitor and enhance human user skills and safety. The initial phase of smart hand tool development entails the seamless integration of sensors and onboard intelligence. In this context, ML has the potential to harness sensor data for various purposes, including the identification of ongoing work tasks, facilitating real-time detection of issues related to tool usage, and recognizing potential safety concerns.

Nevertheless, ML algorithms are inherently data-hungry, which presents a formidable challenge in situations where human operators play a crucial role in data acquisition. This issue is further exacerbated when considering the myriad ways individuals manipulate, wield, and utilize tools to accomplish identical tasks. Synthetic data has emerged as a promising solution for addressing this data requirement \cite{hittmeir2019utility}. This approach involves using algorithms to generate data that mimics real-world scenarios, subsequently employed for training ML models \cite{nikolenko2021synthetic}. However, a persisting challenge revolves around the generation of synthetic data that faithfully replicates the intricacies of human-generated data.  

When examining current instances of synthetic data implementation within robotics applications, such as pose estimation \cite{josifovski2018object} and  sim-to-real transfer \cite{peng2018sim, zhao2020sim}, it becomes evident that the predominant focus centers on leveraging simulations to generate new data. The key difference in our approach lies in its departure from this trend, as we create synthetic data by utilizing a Yaskawa Robot Arm using a smart hand tool equipped with edge devices (Arduino/RaspberryPi) to collect physical sensor data \cite{janapa2023edge}, rather than relying solely on simulation (see Figure~\ref{fig:teaser}). In this context, our technical contributions to prior work are fourfold:

\figTeaser

\begin{enumerate}
    \item \textbf{Smart Hand Tool Development:} We have equipped a rotary power tool (RPT) with a smart tool module (STM). This module comprises an inertial measurement unit (IMU), a current sensor, and a microphone. The amalgamation of these sensors equips us with data that empowers the discrimination of various tasks performed by the tool. These tasks encompass routing, sanding, engraving, and cutting.

    \item \textbf{Synthetic Data Collection:} We generated synthetic data using a robot arm capable of simulating four commonplace tasks frequently encountered by a human using a rotary power tool (RPT). The dataset generated through our methodology encompasses 11 distinct physical signals, meticulously measured by the smart tool module (STM).    

    \item \textbf{Measuring Data Efficacy:} To quantitatively assess the efficacy of the proposed data-centric approach, we conduct a comparative analysis. This evaluation involves contrasting the performance of zero-shot training, using data collected from the smart hand tool operated by humans against the fine-tuning of a model pre-trained using synthetic data obtained from the robot arm equipped with the RPT. These experiments are discussed in detail in Section~\ref{sec:Experimental-Results}.

    \item \textbf{Open-Source Contribution:} We have made available approximately 20 hours of data that encompasses recordings collected from the rotary power tool by humans and the robot arm. In addition, pre-trained models using the synthetic data generated by the Yaskawa robot are also available at: \href{https://github.com/UTAustin-SwarmLab/Smart-Tools}{https://github.com/UTAustin-SwarmLab/Smart-Tools}

\end{enumerate}

\section{Related Work}


Humans have continuously evolved hand tool technology, and this work lends to ongoing efforts \cite{Zoran2014wisechisel,Bendana2023}, notably taking advantage of sensing and machine learning innovations. While the intent is to leave the tool in the hand of the human, robotic partners are essential. Past and present works have examined how robots can be trained to manipulate tools, in many cases by learning from humans \cite{LiuAsada1991skillstorobots,LiuAsada1993deburring,Holladay2019planningtooluse,Shirai2023tactiletoolmanipulation}. Indeed, the latter is very complicated, as humans excel in manipulation tasks where tools interact with the environment. In contrast to cited works, the robot used here is focused on replicating basic work tasks in an open-loop programmatic fashion, as will be described. This is found to be sufficient for the purposes of generating synthetic data for this preliminary study. Robots have been used for data collection and synthetic data generation for over a decade. Some of these applications involve perception \cite{li2019connecting}, data augmentation \cite{young2021visual}, soft robotics to mimic human dexterous motion \cite{wang2022deepclaw}, and more. We also distinguish our work from human-robot collaborative tool use, such as by \cite{Tian2021adroid}. Our work tries to harness robotic data collection capabilities by extending it towards general human tool use. 

Monitoring sensors from tools and processes is not unlike the trends seen in smart manufacturing, where streaming sensor data is enabled by Internet-of-Things (IoT) devices \cite{longo2017smart}. Such data has been used to monitor machine health, detect anomalies, and even adjust process parameters to optimize yield. Smart manufacturing has partly been motivated by the demand for customized products in small batches \cite{wang2016towards}, such as in 3D-printing, or to improve sustainability of production schemes \cite{alkaya2015adaptation}. In addition, the need to support humans in work is essential in tasks related to construction, maintenance, and other applications where robotic solutions may not be available. Lastly, prior work \cite{collier2023ISTA23} examines the needs and implications through a broader socio-technical context.


\section{System Architecture} \label{sec:SysArch}

\figSysArch

\subsection{Smart Tool Module} \label{sec:SmartHandTool}

The apparatus developed in this work serves as a platform for demonstrating real-world applications of a smart tool concept. The intent is to identify best practices in smart hand tool design for a given work application. Two prototype sensor modules were used in this work, both composed of a 9-axis inertial measurement unit (IMU), a microphone, and a current sensor. When the tool is in operation, the sensor data is recorded and analyzed, one module by a RaspberryPi microcomputer, the other by an Arduino Nano BLE sense. One prototype, a smart tool module (STM), had a small OLED screen embedded, for communicating with the operator, and a custom-made encasing to protect the sensors and computer from the environment. The IMU, microphone, OLED screen, and 5-way switch are mounted to the STM and placed in a 3D-printed box encased in aluminum. The setup can be replicated with low-effort soldering and manufacturing skills.

A rotary power tool (RPT) was chosen as the platform for testing the smart tool concept (Model 4300AC, Dremel, Racine, WI, USA). A typical RPT is shown in Figure~\ref{fig:teaser}. This hand-held power tool has an electrical AC motor directly coupled to a drive-shaft that can be manually adjusted to rotate at a desired speed (5,000 to 35,000 revolutions per minute, or rpm) by a user. Different tool bits can be attached to the drive-shaft to change the tool function. A RPT is not typically used in applications requiring high levels of applied force. Rather, a user guides the tool, controlling how the tool interacts with a workpiece. A wide range of attachments enable a RPT to be used for different purposes such as polishing, grinding, cutting, and routing. This versatility and the diverse manual control actions required by a human user are key reasons for selecting a RPT for this preliminary study in smart tool development.

\subsection{Yaskawa Robot}\label{sec:YaskawaRobot}

\figSTM

To accomplish the tasks set out in Section~\ref{sec:Experimental-Results} and produce synthetic data, a programmable automatic machine is needed. In this case, we selected the Yaskawa SDA10D robot due to its ability to execute movements similar to those of a human using a hand tool. The Yaskawa SDA10D robot is an industrial robot that is equipped with two robotic arms that operate independently. The robot features dual 7-axis arms that enable it to exhibit human-like flexibility when performing a variety of movements. Moreover, its programming environment is highly adaptable, as it provides the ability to modify not only position values but also speed and acceleration. The robot can be controlled by manipulating each of its joints individually, or alternatively, by regulating the XYZ position of the end-effector.

Prior to programming the movement patterns and constructing the tool-fastening system to the end-effector of the robot arm, the variables involved in routing, sanding, engraving, and cutting processes that a human would undertake were examined. Human execution of these tasks is not consistent, as it is a stochastic process due to varying compliance of a user's arm as it reacts to the resistance of the material being machined. In addition, the anisotropic material of wood, as used for these experiments, also contributes to variability. Furthermore, factors such as speed, trajectory, working depth, working and travel angles generated by humans are subject to variation. 

\subsection{Data Collection}\label{sec:DataCollection}
Our software stack comprises three main components: data collection, data cleaning, and model building. The data collection process is facilitated by an Arduino microcontroller that communicates with the sensor module on the RPT. The Arduino is responsible for collecting dataframes in the form of CSV files containing the sensor readings. The data collection process was carried out using two methods: the first involved human subjects performing the tasks, while the second utilized the Yaskawa Robot, described as follows.

The same RPT, sensor suite, and experimental setup used during human data collection was mounted on the robotic arm for data collection purposes. To imitate human user use and operating variability, and owing to the fact that robotic movements are more uniform, a compliant robotic arm-tool attachment was developed. Passive compliance, introduced by a rubber insert as shown in Figure~\ref{fig:STM}, introduces variations in trajectory, compliance, work angles and travel angles. Variability in working depth and speed was introduced by programming of the robot.

Movement patterns for each activity were programmed and executed using Cartesian control based on the coordinates of the robot arm's end-effector. Specifically, XYZ coordinates were determined to simulate human-like movements of routing, sanding, engraving, and cutting with a RPT tool. The working depth was programmed based on the maximum and minimum values observed during human performance using the RPT, with the depth value being adjusted with each tool pass. A trapezoidal velocity profile was selected for the Cartesian control programming. Combining these parameters allowed for a better dispersion of data that is comparable to that produced by a human user. All of the programming was carried out using Robot Operating System (ROS1), on version Noetic Ninjemys (ROS Noetic) in C++. The movement patterns were encoded into a 6-DOF vector for each task. 

Once data was collected, we performed data cleaning to remove outliers. The human-collected data was found to be relatively ``dirtier" than the robot-collected data, and we removed about 30\% of the human data after outlier removal. In contrast, the robot-collected data was cleaner as the robot was programmed to perform the tasks consistently for every run. Only about 10\% of robot-collected data was removed after discounting outliers.

Consenting volunteers were instructed on how to use the instrumented RPT in the four activities: cutting, engraving, routing, and sanding, with the bits shown in Figure \ref{fig:ToolBits}. A high-level verbal description with minimal instructions were given for each activity as follows.

\figToolBits

\textbf{Cutting.} The 1.5 inch wood cutting wheel is installed on the tool and a 0.5 inch diameter round wooden dowel rod is secured to a workbench. The user is asked to slice approximately 1/4 inch discs over a time period.

\textbf{Engraving.} The engraving bit is installed on the rotary tool and a piece of standard `2-by-4' lumber measuring 12 inches in length was secured to a workbench. Each user was asked to engrave numbers from zero through nine continuously over a time period.

\textbf{Routing.} The straight routing bit is installed on the rotary tool and a piece of standard `2-by-4' lumber measuring 12 inches in length was secured to a workbench. Each user routes approximately one-eighth inch deep grooves across the width of the wood over a designated time period.

\textbf{Sanding.} The sanding drum is installed on the rotary tool. The subject creates a one half inch \emph{chamfer} on every edge of the `2-by-4' lumber over a designated time period.
    
The participants were given instructions on how to set-up the different tool bits. Some pointers on how to use the tool for a given task were provided, but users were given freedom to achieve each task based on their own comfort with the tool, experience, and unique style. This approach was taken to allow for variability in how each person operated the tool and completed the task.
    
Data collection was conducted for about a one and a half month period. A total of seven participants (five graduate students and two faculty) used the instrumented RPT to work on wooden samples. Each participant performed nine runs of each of the 4 tasks. Each sample run was set for three minutes. A total of 204 sample runs or 612 minutes of data was collected. The nine runs were broken down into 3 runs of training data, 3 runs of validation data, and 3 runs of testing data. After each test, the `2-by-4' wood samples were collected and organized for future analysis. In addition, each run was video recorded for future research work using images and raw sound. 

\figRawSensorDataRouting
 
After collecting data from each subject performing the four different activities, three CSV files were concatenated into three different data frames containing the raw data for train, validate, and test data. These raw data frames were then preprocessed to produce offline training data for the ML models. Data pre-processing involved breaking down the data into 10 second frames with 50\% overlap. The frame size was selected through an iterative process which
resulted in 10 seconds producing the best results, while the 50\% overlap was chosen to feed the model more data. Ten statistical features (minimum, maximum, kurtosis, standard error of the mean, standard deviation, variance, sample skewness, median absolute deviation, and sum) where extracted from each sensor measurement (9 axis IMU, microphone, current sensor) for a total of 110 features. Subsequently, the data was normalized to values from 0 to 1.

\section{Experimental Results}\label{sec:Experimental-Results}

In this section, we present results obtained from our study that aim to answer three key hypotheses. Firstly, we compare the distributions of raw sensor data values when collected for humans versus the Yaskawa robot. Secondly, we compare the performance of zero-shot training on data collected from the tool used by humans with the fine-tuning of a pre-trained model trained on synthetic data obtained from the Yaskawa robot. Lastly, we investigate fine-tuning on individual subjects using the pre-trained model on synthetic data. The next paragraphs discuss the three motives, respectively.



\subsection{\textbf{Hypothesis 1: Data from the robot can be effectively used for pre-training an ML model.}}

Firstly, we compare typical experimental realizations of data collected during robot and human-use in Figure~\ref{fig:RawDataRouting}. It is clear that, although the Yaskawa robot's movements were sharper and more consistent, they provide a baseline for signal features one can observe from during these activities. As the human subjects pick up the tool to perform a routing pass and proceed to move the tool laterally along the lumber repetitively, a candid periodicity is observed which the robot neatly replicates. This empirical confirmation successfully alluded to using the robot in our ML training pipeline. 


\figSensorComparisons

Secondly, we investigate the distributions of the raw sensor values collected from our smart hand tool using human subjects as well as the Yaskawa robot. It is crucial to ensure that the distributions of both datasets are similar yet possess variations to account for real-world data distribution shifts.

To address this, we use a grouped boxplot to examine each sensor's raw value ranges. Each sensor is represented by two boxes that compare the robot-collected and the human-collected value distribution. This is shown in Figure~\ref{fig:SensorComparisons}. Additionally, the diamonds in each x-axis entry represent the outliers of that distribution. This analysis enables us to gain insights into the similarities and differences between the human and robot-collected data. It shows that the robot and human data are roughly similar in variation, accounting for stochasticity of human use and wood heterogeneity. This similarity is essential for using robot-collected data to pre-train our classification model. 

\subsection{\textbf{Hypothesis 2: Pre-training the model will help generalize well for human-collected data.}}

\figZSPreTrain

Now, we focus on demonstrating the practical application of our approach. Most distributed ML systems in production generally house a large model in the server, and fine-tune local models for each specific client. With the data collected by the Yaskawa robot, we can train a large model and fine-tune it on human-collected data. Note that `large' in this context means lots of data; the local models for each user will only be trained on a tool user's collected data. 

In this case, we experimented on all human-collected data. Subsequently, we perform the same analysis for individual human subject-collected data. To systematically evaluate the effectiveness of our approach, we train the model on varying percentages (x\%) of human data in two ways: (1) \emph{zero-shot} training, where we train a new model simply on human-collected data, and (2) pre-trained fine-tuning, where we use a checkpoint of a model trained on all Yaskawa robot data. We use test set accuracy for evaluation, since this is multi-class classification. Test accuracy measures the accuracy of the model on unseen test data. The results are presented in Figure~\ref{fig:ZSPretrainAllHumans}, which compares the two approaches across different percentages of training data.

\subsection{\textbf{Hypothesis 3: Pre-training the model quantifiably improves performance for individual subjects.}}

\figZSPreTrainSubjects


Finally, we fine-tune models for individual subjects. Test data for fine-tuning is only drawn from human-collected data that is isolated from the training pipeline. We conducted In-Distribution (ID) and Out-of-Distribution (OoD) tests. Separating each human subject's data at the start and creating a training pipeline for all human data except for the subject in question is Out-of-Distribution testing. Using all human data including the subject in question for training and then testing on all subjects is testing In-Distribution.

The goal of this experiment is to determine if (a) pre-training works to boost accuracy, and (b) In-Distribution is better than Out-of-Distribution training and testing. The results are presented in Figure~\ref{fig:ZSPreTrainSubjects}, which is a bar plot comparing both OoD and ID testing. These results show that the green bars denoting pre-trained fine-tuning on In-Distribution data works best for our scale of models. Our key takeaway is that, by virtue of pre-training on robot data, accuracy on unseen human test data is improved by 11.4\% on average (and as high as 36\%), although this isn't consistent in some subjects (subjects 5 and 6). 

\subsection{Limitations and Future Work}

Our preliminary work shows that synthetic data can be a viable solution for bridging the data requirement gap in training ML models for human work task recognition. While our experiments show promise for scalability, a truer benchmark of scale is required using many more smart tools, Another key step is to investigate the data collection and ML model deployment `on tool' using edge devices. This will be critical for online feedback and active assistance to a tool user. Lastly, human work tasks require ways to track quality of work. Some preliminary ideas for assessing quality of work for the RPT tasks investigated here are described in \cite{Bendana2023}. Future researchers are encouraged to use data we collected in \ref{sec:DataCollection}, possibly from camera data of the `2-by-4' lumber to study quality of work.

\vspace{-2mm}

\section{Conclusion}

We have presented a way for combining human and robot generated synthetic data to train a prototype smart hand tool intended for human use. Our experimental results show the effectiveness and scalability of fine-tuning machine learning models that can identify typical work tasks. However, there are limitations to our approach, such as the need for a model zoo to perform model selection and improve individual user's experience, testing on a larger dataset of human subjects, and testing for quality of work. Overall, the proposed approach provides a promising direction for showing how synthetic data can be collected and used for smart hand tool development.


\section*{Acknowledgements}
This work was supported by funding from Good Systems, a research grand challenge at the University of Texas at Austin, Microsoft Research, and The MITRE Corporation. We thank Dr. Ashish D. Deshpande for providing access to the Yaskawa SDA10D robot for our synthetic data collection process. Elena Soto's insight and assistance in programming the Yaskawa Robot for the synthetic data collection process is greatly appreciated. Finally, thanks to Burzin Balsara, Elena Soto, Jose Bendana, Kathy Hill, Pablo Pejlatowicz, Raul G. Longoria, and Sandeep Chinchali for manually performing tasks and helping collect many hours of human data. 

{\small
\bibliographystyle{unsrt}
\bibliography{bibs.bib}

\begin{thebibliography}{10}

\bibitem{hittmeir2019utility}
Markus Hittmeir, Andreas Ekelhart, and Rudolf Mayer.
\newblock On the utility of synthetic data: An empirical evaluation on machine
  learning tasks.
\newblock In {\em Proceedings of the 14th International Conference on
  Availability, Reliability and Security}, pages 1--6, 2019.

\bibitem{nikolenko2021synthetic}
Sergey~I Nikolenko.
\newblock {\em Synthetic data for deep learning}, volume 174.
\newblock Springer, 2021.

\bibitem{josifovski2018object}
Josip Josifovski, Matthias Kerzel, Christoph Pregizer, Lukas Posniak, and
  Stefan Wermter.
\newblock Object detection and pose estimation based on convolutional neural
  networks trained with synthetic data.
\newblock In {\em 2018 IEEE/RSJ international conference on intelligent robots
  and systems (IROS)}, pages 6269--6276. IEEE, 2018.

\bibitem{peng2018sim}
Xue~Bin Peng, Marcin Andrychowicz, Wojciech Zaremba, and Pieter Abbeel.
\newblock Sim-to-real transfer of robotic control with dynamics randomization.
\newblock In {\em 2018 IEEE international conference on robotics and automation
  (ICRA)}, pages 3803--3810. IEEE, 2018.

\bibitem{zhao2020sim}
Wenshuai Zhao, Jorge~Pe{\~n}a Queralta, and Tomi Westerlund.
\newblock Sim-to-real transfer in deep reinforcement learning for robotics: a
  survey.
\newblock In {\em 2020 IEEE symposium series on computational intelligence
  (SSCI)}, pages 737--744. IEEE, 2020.

\bibitem{janapa2023edge}
Vijay Janapa~Reddi, Alexander Elium, Shawn Hymel, David Tischler, Daniel
  Situnayake, Carl Ward, Louis Moreau, Jenny Plunkett, Matthew Kelcey, Mathijs
  Baaijens, et~al.
\newblock {Edge Impulse: An MLOps Platform for Tiny Machine Learning}.
\newblock {\em Proceedings of Machine Learning and Systems}, 5, 2023.

\bibitem{Zoran2014wisechisel}
Amit Zoran, Roy Shilkrot, Pragun Goyal, Pattie Maes, and Joseph~A. Paradiso.
\newblock The wise chisel: The rise of the smart handheld tool.
\newblock {\em IEEE Pervasive Computing}, 13(3):48--57, 2014.

\bibitem{Bendana2023}
Jose Bendana.
\newblock Analyzing power driven hand tool operation using low-cost sensors.
\newblock Master's thesis, The University of Texas at Austin, Mechanical
  Engineering Department, 2023.

\bibitem{LiuAsada1991skillstorobots}
Sheng Liu and Haruhiko Asada.
\newblock {Transferring Manipulative Skills to Robots: Representation and
  Acquisition of Tool Manipulative Skills Using a Process Dynamics Model}.
\newblock {\em Journal of Dynamic Systems, Measurement, and Control},
  114(2):220--228, 06 1992.

\bibitem{LiuAsada1993deburring}
S.~Liu and H.~Asada.
\newblock Teaching and learning of deburring robots using neural networks.
\newblock In {\em [1993] Proceedings IEEE International Conference on Robotics
  and Automation}, pages 339--345 vol.3, 1993.

\bibitem{Holladay2019planningtooluse}
Rachel Holladay, Tomás Lozano-Pérez, and Alberto Rodriguez.
\newblock Force-and-motion constrained planning for tool use.
\newblock In {\em 2019 IEEE/RSJ International Conference on Intelligent Robots
  and Systems (IROS)}, pages 7409--7416, 2019.

\bibitem{Shirai2023tactiletoolmanipulation}
Yuki Shirai, Devesh~K. Jha, Arvind~U. Raghunathan, and Dennis Hong.
\newblock Tactile tool manipulation.
\newblock In {\em 2023 IEEE International Conference on Robotics and Automation
  (ICRA)}, pages 12597--12603, 2023.

\bibitem{li2019connecting}
Yunzhu Li, Jun-Yan Zhu, Russ Tedrake, and Antonio Torralba.
\newblock Connecting touch and vision via cross-modal prediction.
\newblock In {\em Proceedings of the IEEE/CVF Conference on Computer Vision and
  Pattern Recognition}, pages 10609--10618, 2019.

\bibitem{young2021visual}
Sarah Young, Dhiraj Gandhi, Shubham Tulsiani, Abhinav Gupta, Pieter Abbeel, and
  Lerrel Pinto.
\newblock Visual imitation made easy.
\newblock In {\em Conference on Robot Learning}, pages 1992--2005. PMLR, 2021.

\bibitem{wang2022deepclaw}
Haokun Wang, Xiaobo Liu, Nuofan Qiu, Ning Guo, Fang Wan, and Chaoyang Song.
\newblock Deepclaw 2.0: A data collection platform for learning human
  manipulation.
\newblock {\em Frontiers in Robotics and AI}, page~38, 2022.

\bibitem{Tian2021adroid}
Rundong Tian and Eric Paulos.
\newblock Adroid: Augmenting hands-on making with a collaborative robot.
\newblock In {\em The 34th Annual ACM Symposium on User Interface Software and
  Technology}, page 270–281, New York, NY, USA, 2021. Association for
  Computing Machinery.

\bibitem{longo2017smart}
Francesco Longo, Letizia Nicoletti, and Antonio Padovano.
\newblock Smart operators in industry 4.0: A human-centered approach to enhance
  operators’ capabilities and competencies within the new smart factory
  context.
\newblock {\em Computers \& industrial engineering}, 113:144--159, 2017.

\bibitem{wang2016towards}
Shiyong Wang, Jiafu Wan, Daqiang Zhang, Di~Li, and Chunhua Zhang.
\newblock Towards smart factory for industry 4.0: a self-organized multi-agent
  system with big data based feedback and coordination.
\newblock {\em Computer networks}, 101:158--168, 2016.

\bibitem{alkaya2015adaptation}
Emrah Alkaya, Merve Bogurcu, Ferda Ulutas, and G{\"o}ksel~Niyazi Demirer.
\newblock Adaptation to climate change in industry: Improving resource
  efficiency through sustainable production applications.
\newblock {\em Water Environment Research}, 87(1):14--25, 2015.

\bibitem{collier2023ISTA23}
Chelsea Collier, Kenneth~R. Fleischmann, Tinna Lassiter, Sherri~R. Greenberg,
  Raul~G. Longoria, and Sandeep Chinchali.
\newblock Co-designing socio-technical interventions with skilled trade
  workers.
\newblock In {\em Proceedings of the IEEE International Symposium on Technology
  and Society (ISTAS23), Public Interest Technology University Network}, 2023.

\end{thebibliography}
}

\end{document}